# SIMULATION OPTIMIZATION OF THE CROSSDOCK DOOR ASSIGNMENT PROBLEM

Dr. Uwe Aickelin
Adrian Adewunmi
ASAP, School of Computer Science and IT, University of Nottingham, Jubilee Campus,
Wollaton Road, Nottingham, NG8 1BB, UK.
uwe.aickelin@nottingham.ac.uk, aqa@cs.nott.ac.uk

**ABSTRACT:**

*The purpose of this report is to present the Crossdock Door Assignment Problem, which involves assigning destinations to outbound dock doors of Crossdock centres such that travel distance by material handling equipment is minimized. We propose a two fold solution; simulation and optimization of the simulation model – simulation optimization. The novel aspect of our solution approach is that we intend to use simulation to derive a more realistic objective function and use Memetic algorithms to find an optimal solution. The main advantage of using Memetic algorithms is that it combines a local search with Genetic Algorithms. The Crossdock Door Assignment Problem is a new domain application to Memetic Algorithms and it is yet unknown how it will perform.*

Keywords: Crossdock Door Assignment Problem, Discrete Event Simulation, Optimization, Genetic Algorithms.

## 1. INTRODUCTION

Traditionally, warehouses have had the following functions: receiving, storage, order picking and shipping. Logistics companies have found storage and order picking to be cost intensive, this lead to a strategy of keeping zero inventories. This strategy based on the Just in Time philosophy eliminates the storage and order picking function of the warehouse while maintaining the receiving and shipping activity [3]. Presently, each incoming trailer is assigned an available inbound door as soon as it arrives and each outbound trailer is assigned a specific single outbound door. In addition to inbound and outbound doors, open doors are available for frequent reassignment. The arrangement of inbound and outbound doors and the assignment of destinations to outbound doors is a difficult combinatorial optimization problem. This is due to the dynamic nature of trailer scheduling, flow of freight pattern through the Crossdock centre, the mix of freight that has to be bulk broken and consolidated and material handling costs. The efficiency of the Crossdock centre is enhanced by obtaining optimal assignment and arrangement of doors. Bartholdi and Gue report an implementation of optimal dock door assignment increased productivity by more than 11% [1]. The Crossdock Door Assignment Problem is a problem related to the Dock Door Assignment Problem, first formulated by Tsui and Chang [7]. The Crossdock Door Assignment Problem's objective is to find an optimal arrangement of a Crossdock centre's inbound and outbound doors and the most efficient assignment of destinations to outbound doors, such that distance travelled by material handling equipment is minimized. A good solution to this problem will be to make Crossdock Door Assignment dynamic and proactive enough to respond quickly to changes in freight flow patterns thus reducing material handling costs and increase the amount of freight processed through the Crossdock centre.

An illustrative example of the Crossdocking centre is presented below:

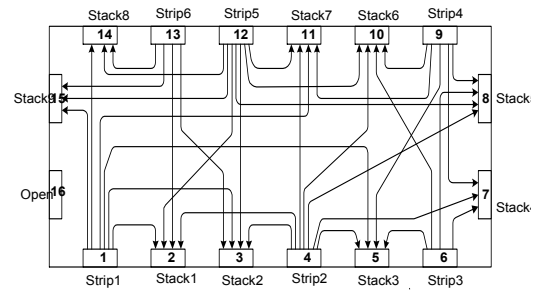

Figure 1. Example Crossdocking centre with 16 doors; 6 strip doors (Inbound Doors), 9 stack doors (Outbound Doors), and 1 open door (no assignment). Ref: [2]

There have been a number of proposed solutions to the Crossdock Door Assignment Problem e.g. [7] and [1] focused on using local search technique, [2] implemented a metaheuristic approach using Genetic Algorithms and [5] utilized simulation techniques. The aim of this project is to develop a robust Metaheuristic to solve the Crossdock Door Assignment Problem where the objectives include; increasing the efficiency and throughput of the Crossdock centre, reduce material handling costs including minimizing travelling distances for material handling equipment

## 2. CROSSDOCK DOOR ASSIGNMENT PROBLEM

The Crossdock door Assignment Problem is related to the Dock Door Assignment Problem, first formulated by Tsui and Chang [7]. The Crossdock Door Assignment Problem objective is to find the optimal arrangement of a Crossdock centre's inbound and outbound doors and the most efficient assignment of destinations to outbound doors, such that the distance travelled by material handling equipment is minimized. It is assumed that there are $I$ inbound doors, $J$ outbound doors, $M$ origins and $N$ destinations for the Crossdock centre, $I \geq M$ and $J \geq N$. Let $X_{im} = 1$ if origin $m$ is assigned to inbound door $i$, $X_{im} = 0$ otherwise. Let $Y_{nj} = 1$ if destination $n$ is assigned to outbound door $j$, $Y_{nj} = 0$ otherwise. Let $d_{ij}$ represent the distance between inbound door $i$ and outbound door $j$. Let $w_{mn}$ represent the number of trips required by the material handling equipment to move items originating from m to the Crossdock door where freight destined for n is being consolidated. This section defines a mathematical model for the Crossdock door Assignment Problem based on work by [7] and [2]. It is formulated below:

### 2.2 MATHEMATICAL MODEL

#### 2.2.1 PARAMETERS

M - Number of $M$ origins.
N - Number of $N$ destinations.
I - Number of $I$ inbound doors.
J - Number of $J$ outbound doors.
$w_{mn}$ - represent the number of trips required by the material handling equipment to move items originating from m to the Crossdock door where freight destined for n is being consolidated.
$d_{ij}$ - Represents the distance between inbound door $i$ and outbound door $j$.

#### 2.2.2 DECISION VARIABLES

$X_{im}$ - $X_{im} = 1$ if origin $m$ is assigned to inbound door $i$, $X_{im} = 0$ otherwise.
$Y_{nj}$ - $Y_{nj} = 1$ if destination $n$ is assigned to outbound door $j$, $Y_{nj} = 0$ otherwise.

#### 2.2.3 OBJECTIVE FUNCTION

Minimizing

$$\sum_{i=1}^{I} \sum_{j=1}^{J} \sum_{m=1}^{M} \sum_{n=1}^{N} d_{ij} w_{mn} X_{mi} Y_{nj}$$

## 3. PROPOSED PLAN OF WORK

### 3.1 DISCRETE EVENT SIMULATION

In the field of Operational Research, simulation is defined as experimentation with a simplified reproduction of a real life model (on a computer) as it progresses through time, for the purpose of better understanding and/or improving that system. In the initial stage of our project, a simulation of Crossdock Door Assignment will be performed with the aim of getting a better understanding of the assignment of doors. The simulation will imitate the performance of the Crossdock centre in a controlled environment in order to estimate what the actual performance will be, taking parameter changes like Crossdock shape, inbound and outbound door number, trailer schedule as well as freight flow and mix over a period in time into consideration. The type of simulation that is applicable to Crossdock Door Assignment Problem is Discrete Event Simulation, where changes to the system occur immediately at arbitrary points in time as a result of the occurrence of discrete events. The characteristic of discreteness best suits our problem because each inbound trailer joins a queue to compete for the scare resource of inbound door and once freight is bulk broken it competes for the scare resource of outbound door. In particular, we will simulate the flow of freight between inbound and outbound doors and as well as the arrangement of the Crossdock doors. The quality of the scenario will be measured by the overall cost of this flow of freight; this represents a refinement of the objective function (1). This is because the objective function (1) will take into account issues such as bottleneck and delays.

### 3.2 SIMULATION OPTIMIZATION

A current trend in simulation is the idea of optimizing the results of simulation models. The idea is search for an optimal solution for simulation models with as many decision variables as you desire. This poses the problem of noisy objective functions, which we will investigate with respect to the Crossdock Door Assignment Problem. [6] present two simulated annealing algorithms that are designed to handle noisy objective functions;

we are interested in comparing the performance of Memetic algorithms to other popular heuristics in relation to noisy objective functions [4].

After simulating the different possible Crossdock Door Assignments, the next task is to find the best door to freight and trailer to door assignment. This can be achieved by optimizing the door assignment from the simulation models that performed the best against predetermined criteria. As iterated, the purpose of this research is to present a novel search method for solving the Crossdock Door Assignment Problem which emphasises on exploring global search for promising solutions within the whole feasible region, while exploiting local searches for optimal solutions. We are going to find the optimal door assignment using Memetic Algorithms, with the objective to bring new insights into the application of Memetic algorithm on Crossdock door assignment problem.

## 4. SUMMARY

In this report, we propose to study the Crossdock door Assignment Problem. In contrast to similar problems like the Dock Door Assignment Problem by [7], our Crossdock door Assignment Problem's objective is to find the optimal arrangement of a Crossdock centre's inbound and outbound doors and the most efficient assignment of destinations to outbound doors, such that the distance travelled by material handling equipment is minimized. The solution we put forward is a new methodology for solving the Crossdock door Assignment Problem. It places emphasises on combining simulation and optimisation while exploring the global search space for promising solutions within the whole feasible region, and exploiting local searches for optimal solutions (Memetic Algorithms).

**AUTHORS BIOGRAPHIES:**

**DR. UWE AICKELIN** is a Senior Lecturer in Computer Science at the School of Computer Science and IT, University of Nottingham and a member of Automated Scheduling, Optimisation & Planning Research group (ASAP) School of Computer Science and IT, University of Nottingham, Jubilee Campus, Wollaton Road, Nottingham, NG8 1BB, UK..http://www.cs.nott.ac.uk/~uxa/

**ADRIAN ADEWUNMI** received a BSc (Hons) Computer Science from the University of Bradford in 2005. He is currently a PhD student in Computer Science at the School of Computer Science and IT, University of Nottingham and a member of the Automated Scheduling, Optimisation & Planning Research group (ASAP), School of Computer Science and IT, University of Nottingham, Jubilee Campus, Wollaton Road, Nottingham, NG8 1BB, UK.http://www.cs.nott.ac.uk/~aqa/home-asap.shtml